# Hyb-KAN ViT: Hybrid Kolmogorov-Arnold Networks Augmented Vision Transformer

Sainath Dey, Mitul Goswami, *Student Member, IEEE*, Jashika Sethi, and Prasant Kumar Pattnaik, *Senior Member, IEEE*

*Abstract—*. This study addresses the inherent limitations of Multi-Layer Perceptrons (MLPs) in Vision Transformers (ViTs) by introducing Hybrid Kolmogorov-Arnold Network (KAN)-ViT (Hyb-KAN ViT), a novel framework that integrates wavelet-based spectral decomposition and spline-optimized activation functions, prior work has failed to focus on the prebuilt modularity of the ViT architecture and integration of edge detection capabilities of Wavelet functions. We propose two key modules: Efficient-KAN (Eff-KAN), which replaces MLP layers with spline functions and Wavelet-KAN (Wav-KAN), leveraging orthogonal wavelet transforms for multi-resolution feature extraction. These modules are systematically integrated in ViT encoder layers and classification heads to enhance spatial-frequency modeling while mitigating computational bottlenecks. Experiments on ImageNet-1K (Image Recognition), COCO (Object Detection and Instance Segmentation), and ADE20K (Semantic Segmentation) demonstrate state-of-the-art performance with Hyb-KAN ViT. Ablation studies validate the efficacy of wavelet-driven spectral priors in segmentation and spline-based efficiency in detection tasks. The framework establishes a new paradigm for balancing parameter efficiency and multi-scale representation in vision architectures.

*Index Terms*—Vision Transformers, Kolmogorov Arnold Networks, Wavelet functions, Image Recognition, Object Detection, Semantic Segementation

## Introduction

Vision Transformers (ViTs) have revolutionized the field of computer vision by introducing a novel approach to image analysis that departs from traditional convolutional neural networks (CNNs). A. Dosovitskiy et al. modified the original transformer architecture to demonstrate its capability in vision tasks [1][2], by leveraging the self-attention mechanism as demonstrated by M. T. Loung et al., which was originally developed for natural language processing [3]. Central to their design is the reliance on Multi-Layer Perceptrons (MLPs), which serve as critical components for both feature extraction within transformer layers and final classification tasks. In the classification head, MLPs take the final encoded representation, particularly the [CLS] token, and map it to task-specific outputs such as class probabilities.

X. Yang et al. introduced the idea of Kolmogorov–Arnold Transformers (KATs) that extend the ViT framework by integrating Kolmogorov-Arnold Networks (KANs) with rational functions to enhance feature encoding and classification [4]. KANs, inspired by functional approximation theory, decompose complex multivariate mappings into simpler univariate transformations, offering a novel computational paradigm for efficient representation learning [5]. An emerging modification to KAN, known as Wavelet-KAN proposed by Z. Bozorgasl et al., efficiently captures both high-frequency and low-frequency components of input data, leveraging wavelet-based multi-resolution analysis for robust feature extraction [6]. Unlike traditional KANs, which often struggle with overfitting or computational inefficiency, Wav-KAN employs orthogonal or semi-orthogonal wavelet bases to balance accurate data representation with noise reduction. While KANs and wavelet transformations have shown promise individually, their integration into ViT architecture remains unexplored.

To systematically explore the integration of Wav-KAN into ViTs, the authors developed a modular framework that replaces traditional MLPs with altered Wav-KAN modules in both encoder layers and classification heads. This modular design allows flexibility in experimenting with different configurations, enabling a detailed examination of how Wav-KAN enhances feature encoding and representation learning within ViTs [2]. The motivation behind replacing MLPs in ViTs with Wav-KAN stems from the limitations of MLPs. Wav-KAN addresses the challenges, by leveraging wavelet transformations and KAN, enabling improved capacity to model complex data patterns within transformer architectures [6]. Further, our work introduces Hybrid Modules that integrate Wav-KAN and Efficient-KAN (Eff-KAN) architectures within the ViT framework. These modules are strategically deployed in both encoder layers and classification heads to enhance feature representation and computational efficiency. The Eff-KAN leverages spline-based activation functions, enable smoother, adaptive decision boundaries through approximations maintaining the capacity to model complex data patterns. On the other hand, Wavelet-KAN incorporates wavelet-based multi-resolution analysis, enabling robust extraction of both high-frequency and low-frequency components.

Section II reviews CNNs and ViTs, contextualizing backbone architecture evolution. Section III details the framework design, integrating MLP-based transformers with spline-optimized and wavelet-KAN modules to overcome representation limits. Section IV outlines experiments, including configurations, dataset selection, and validation protocols. Section V evaluates the framework via ablation studies and benchmarking, quantifying accuracy and efficiency gains. Section VI concludes by synthesizing the hybrid KAN-transformer contributions and proposing future adaptive vision research directions.

## II. RELATED WORKS

*2.1. Convolutional Neural Network Backbones*

CNNs have been central to computer vision research, particularly since AlexNet's breakthrough [7]. Early approaches concentrated on deepening networks to boost their performance, as seen with VGG [8], which utilized a straightforward yet effective design with consistent 3×3 convolutional kernels. The development of ResNet [9] represented a pivotal innovation by introducing skip connections. GoogLeNet [10] and InceptionNet [11] pioneered a split-transform-merge approach by incorporating multiple kernel paths into a single CNN block. ResNeXt [12] extended this idea by developing a uniform multi-path architecture, demonstrating that increasing the number of parallel paths—cardinality—can be more effective than simply expanding width or depth. DenseNet [13] introduced a dense connectivity scheme where each layer receives feature maps from all preceding layers. Moreover, SparseNet [14] refined this concept by spacing the dense connections exponentially, further optimizing parameter utilization.

To enhance computational efficiency for mobile and edge devices, architectures such as ShuffleNet [15] were introduced. ShuffleNet employs point-wise group convolution along with channel shuffle operations, effectively reducing computational complexity while maintaining high accuracy [9]. MobileNet [16] adopted depth-wise separable convolutions to build lightweight models tailored for mobile applications, whereas EfficientNet [17] implemented compound scaling to systematically balance network depth, width, and resolution. CSPNet [18] then emerged as a method to bolster CNN learning capabilities while alleviating computational bottlenecks.

*2.1. Vision Transformer Backbones*

ViTs have transformed computer vision by introducing a fundamentally different approach to image analysis compared to traditional CNNs [7]. A. Dosovitskiy et al. adapted the original transformer architecture to showcase its effectiveness in vision tasks [2], utilizing the self-attention mechanism initially designed for natural language processing [3][19]. A key limitation of the original ViT is its computational complexity, stemming from the quadratic time costs in self-attention and large parameter counts [2]. To address this, researchers have developed various transformer architectures with enhanced efficiency and performance. Swin Transformer [20] introduced a hierarchical structure with shifted windows, limiting self-attention computation to non-overlapping local.

DeiT [24] demonstrated that with appropriate training strategies and strong data augmentation, ViTs can be effectively trained on ImageNet without requiring massive pre-training datasets. ConViT [25] introduced gated positional self-attention to incorporate inductive biases from convolutional operations while preserving the flexibility of self-attention. CaiT [26] and DeepViT [27] explored ways to effectively scale transformers to greater depths, with techniques like class-attention layers and re-attention mechanisms. Hybrid approaches that combine CNN and transformer components have also shown promise. T2T-ViT [28] progressively tokenizes images to reduce token sequence length and model overlapping patch information. RegionViT [29] introduced a regional-to-local attention mechanism to reduce the computational burden of standard self-attention by alternating between regional and local self-attention. Dual-ViT [30] splits self-attention into two distinct pathways: a semantic pathway that efficiently compresses token vectors and a pixel pathway that focuses on learning fine-grained pixel-level details. To address the constraints of edge devices with limited computational power, lightweight transformer models such as MicroViT [31] have been developed. Similarly, MobileViT [32] combines the local processing capability of convolutions with the global interaction of transformers, creating an architecture suitable for mobile devices.

While existing ViT variants improve computational efficiency via hierarchical attention [20]–[27], hybrid architectures [28]–[30], or parameter-efficient designs, they still rely on MLP-based projections with limited multi-scale adaptability. In contrast, our Hyb-KAN ViT replaces MLPs with Wav-KAN modules in the encoder for multi-resolution feature extraction and Eff-KAN modules in the classification head for efficient inference. Unlike prior KAN-based methods [4][6] focused on either splines or wavelets, our approach integrates both: Wav-KAN leverages orthogonal wavelets to preserve edges while suppressing high-frequency noise, and Eff-KAN uses GPU-optimized splines to reduce memory usage.

## III. METHODOLOGY

This section begins with a brief review of the conventional patchification, multi-head self-attention block, and positional embeddings adopted in existing ViTs [2] and analyzes them. We next propose new principled Transformer structures, a novel modular ViT framework, designed to explore the impact of different architectural configurations on performance. By integrating Efficient KAN (see Fig. 2(a)) and Wavelet KAN (see Fig. 2(b)) modules as encoder and classification head variants in Fig. 1, the proposed framework provides a flexible and robust approach for visual recognition tasks.

*3.1. MLP-Based Transformer*

MLP-based transformer architecture utilizes multi-head self-attention mechanisms to capture complex dependencies across input sequences [2]. The general algorithm for it is demonstrated in Algorithm 1.

---

**Algorithm 1** MLP-Based Transformer

**INPUT:**
Feature Vector $x_i \in \mathbb{R}^{n \times d}$
**OUTPUT:**
$x_l^n = LayerNorm1(x_l)$
*For* each head $i$ from 1 to $h$

$q_i = x_l^n \times W_i^Q$
$k_i = x_l^n \times W_i^K$
$v_i = x_l^n \times W_i^V$

$$head_i = softmax(\frac{q_i \times k_i^t}{\sqrt{d_h}})v_i$$
$$MHA_{Output} = Concat(head_1 \ldots. head_h) \times W^O$$
$$x'_l = MHA_{Output} + x_l$$
$$x'^n_l = LayerNorm2(x'_l)$$
$$FFN_{Output} = FFN(x'^n_l)$$
$$x_{(l+1)} = FFN_{Output} + x'_l$$
$$return\ x_{l+1}$$

The algorithm processes input features $x_i \in \mathbb{R}^{n \times d}$, where $n$ is the number of tokens, and $d$ is the embedding dimension. First, layer normalization (LN) standardizes $xl$ to $x_l^n$. Multi-Head Attention (MHA) computes queries $q_i = x_l^n \times W_i^Q$, keys $k_i = x_l^n \times W_i^K$, and values $v_i = x_l^n \times W_i^V$ for each $head_i$. Attention scores weight values to produce $head_i$, which are concatenated and projected to $MHA_{Output}$. Residual connections ensure gradient flow. The Feed-Forward Network (FFN), typically an MLP, play a critical role. After self-attention captures token relationships, MLPs independently process each token to refine its representation. They typically consist of two fully connected layers with a non-linear activation in between, as expressed by equation (1).

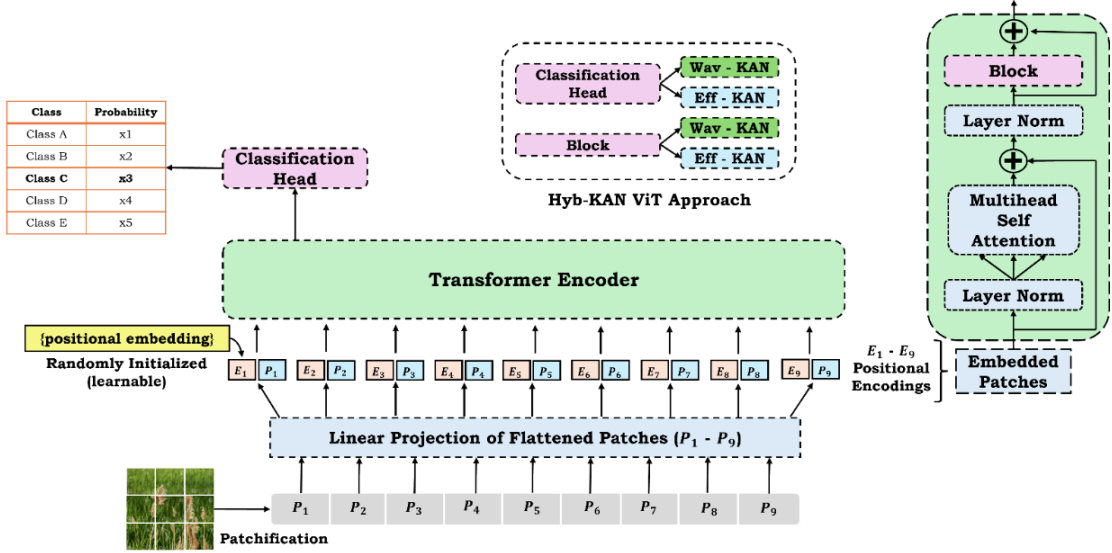

Fig. 1. The illustration of the detailed architecture of our proposed Vision Transformer

$$y = W_2 \times GeLU(W_1 x + b_1) + b_2 \quad (1)$$

The first layer expands the feature dimension, while the second reduces it back, enabling the model to capture complex patterns. However, MLPs have limitations. Existing MLP-based methods, [2], [20], [21], [23], [24], and [25] increase the patch size to reduce the number of tokens, which results in lower-resolution feature maps. Meanwhile, other approaches [22], [26], [27], [29], [30], and [31] confine the attention mechanism within a localized window, thereby limiting computational complexity to a linear scale relative to the input resolution. They process tokens independently, ignoring spatial relationships within patches, and their reliance on linear transformations restricts their ability to model intricate dependencies. This creates scope for replacing MLPs with advanced modules.

*3.2. Efficient KAN-Based Transformer*

KANs offer an alternative to MLPs by replacing fixed activation functions with learnable B-spline functions on network edges. Original KANs [5] faces critical limitations, including excessive memory consumption from expanding intermediate variables, which hinders GPU utilization and slows training [33]. Additionally, suboptimal initialization strategies degraded performance. Eff-KAN [34] resolves these by reformulating spline computations as linear combinations of pre-activated basis functions, enabling GPU-friendly matrix multiplications and adopting Kaiming initialization for both base weights and spline scalers; and replacing input-centric L1 regularization with weight-based regularization, improving scalability and convergence while preserving interpretability.

**Algorithm 2** Efficient-KAN
**INPUT:**
Data sample : $x \in R^n$
Network architecture : $L$ layers
Spline parameters : $k$ knots, grid range $[a, b]$
**OUTPUT:**
$For$ layer $l$ from 1 to $L$
  $W_{base}^{(l)} \in \mathbb{R}^{d_l \times d_{l-1}}$
  $C^{(l)} \in \mathbb{R}^{d_l \times d_{l-1} \times k}$
  $S^{(l)} \in \mathbb{R}^{d_l \times d_{l-1}}$
  $b^{(l)} = \mathbb{R}^d$
Set $x^{(0)} = x$
$For$ layer $l$ from 1 to $L$
  $B^{(l)} \in \mathbb{R}^{n \times d_{l-1} \times k}$ for input $x^{(l-1)}$
  $\phi_{spline}^{(l)} = \left(\sum_{r=1}^{k} C_r^{(l)} \cdot B_r^{(l)}\right) \odot S^{(l)}$

$$z^{(l)} = W_{base}^{(l)} x^{(l-1)} + \phi_{spline}^{(l)} + b^{(l)}$$
$$x^{(l)} = GELU(z^{(l)})$$

$$y = x^L$$
$$return\ y$$

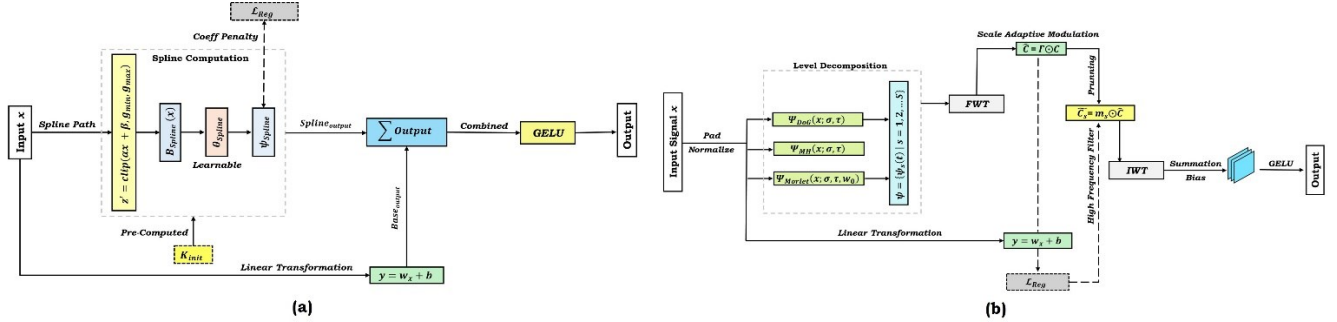

Fig. 2. The illustration of the detailed architecture of the (a) Eff-KAN Block and (b) Wav-KAN Block

In Algorithm 2, $C^{(l)}$, a 3D tensor storing $k$ spline coefficients per connection to d nonlinear curves within the grid range $[a,b]$, $S^{(l)}$, a scaling matrix to dynamically adjust spline outputs, and $b^{(l)}$, a bias vector. Fig. 2(a) illustrates the architecture of the Eff-KAN Block, which utilizes B-spline basis functions along with Linear Transformation, allowing for controlled and sparse activation.

Eff-KAN offers significant advantages over traditional MLPs when deployed in the encoder and classification head of ViTs. In the encoder, where input patches are transformed into high-dimensional representations, KANs excel due to their spline-based activation functions, which adaptively learn smooth, interpretable mappings instead of relying on fixed nonlinearities like ReLU [5]. Replacing MLPs with KANs in the encoder and classification head leverages the mathematical properties of B-spline functions, particularly their local support and smoothness. In the encoder, a two-layer KAN applies B-splines to model edge features. The first KAN layer decomposes input patches into spline terms using equation (2), where $B_k(x; t)$ are B-spline basis functions with knots $t$ and $c_{jk}$ are learnable coefficients.

$$\phi_j^{(1)}(x) = \sum_k c_{jk} B_k(x; t) \qquad (2)$$

These basis functions focus on localized regions like, edges in image patches due to their compact support. The second layer combines these spline terms into higher-order interactions as demonstrated in equation (3), where, $w_{ij}$ adaptively weights edge-relevant features. By composing splines, KANs approximate piecewise smooth functions that align with natural image statistics (sharp edges, gradual textures).

$$\phi_i^{(2)}(\phi^{(1)}) = \sum_j w_{ij} \phi_j^{(1)}(x) \qquad (3)$$

B-splines excel at capturing edges because their knot positions $t$ are trainable. During optimization, knots cluster near regions of high gradient (edges), allowing KANs to allocate resolution where needed. This contrasts with MLPs, which use fixed activation slopes and require more neurons to resolve edges.

In the head, a single KAN layer with B-splines maps the [CLS] token to class scores using equation (4) where, $Z_{CLS}$ is the [CLS] embedding.

$$p(y|x) = Softmax(\sum_k \alpha_k B_k(Z_{CLS}; t)) \qquad (4)$$

The smoothness of B-splines prevents erratic decision boundaries, while their local support avoids overfitting to spurious correlations. B-spline's piecewise polynomial structure and adaptive knots enable KANs to act as "edge detectors" in early layers and "smoothing operators" in later layers. This mimics the human visual system's hierarchical processing, making KANs particularly effective for ViT's encoder (edge-aware feature extraction) and classification head (stable, data-efficient inference).

While KANs demonstrate superior accuracy in ViT encoder and classification head replacements, their adoption introduces significant computational bottlenecks, as highlighted by KAT [4]. The primary drawback lies in their parameter inefficiency, compared to MLPs. For instance, a single KAN layer with n spline bases per activation can demand O(n×d_in×d_out) parameters, whereas an MLP with ReLU needs only O(d_in×d_out) [2][5]. This parameter bloat translates to higher memory usage and slower inference, which is particularly problematic for ViTs, which are already scaled quadratically with token counts. Also, B-spline-based KANs face challenges in GPU efficiency due to their inherently non-parallelizable operations and lack of native CUDA kernel support.

Furthermore, KANs incur higher GFLOPs due to the dynamic evaluation of spline functions, which involve piecewise polynomial calculations and knot-interval searches. Even with the outcomes of an Efficient KAN implementation, they struggle to match MLPs' FLOPs efficiency, as MLP operations map trivially to highly optimized GPU kernels. While KANs' accuracy gains are enticing, their computational demands leave them impractical for large-scale ViT deployments, where low-latency, low-FLOPs inference is critical.

*3.3. Wavelet KAN-Based Transformer*

Unlike B-spline KANs, which rely on unoptimized operations for CUDA kernels, Wav-KANs can leverage pre-optimized wavelet decomposition kernels, reducing latency in frequency-band splitting [35]. However, integrating wavelet bases into trainable KAN layers introduces overheads, such as iterative parameter updates for wavelet coefficients and memory intensive multi-resolution feature storage, which are less GPU-friendly than MLPs' fused matrix multiply-activation pipelines. That said, Wav-KANs are not inherently inefficient—their wavelet operations, when optimized via custom kernels, can approach MLP-level speeds while offering superior interpretability and multi-scale modeling.

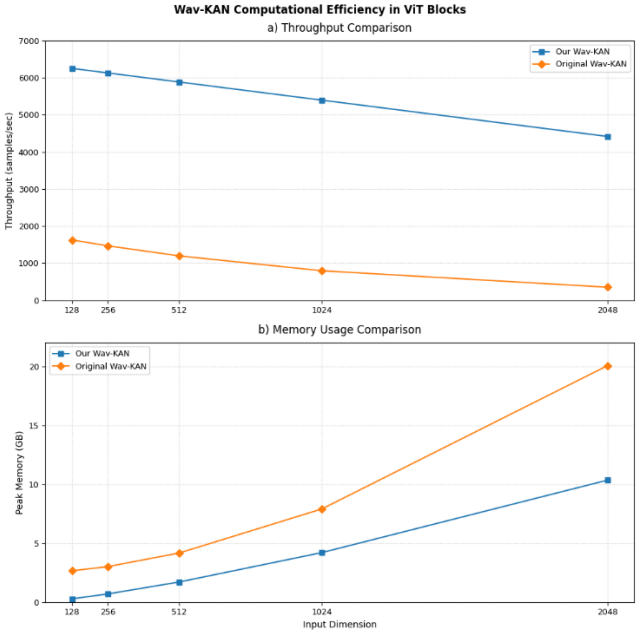

Fig. 3. Comparison of (a) Throughput (sample/s) and (b) Peak Memory (GB) for Input Dimensions Sizes on using the two Wav-KAN implementations in ViT layers.

While existing Wav-KAN implementations [6] leverage continuous (CWT) and discrete wavelet transforms (DWT), their reliance on general-purpose tensor operations introduces significant parameter counts and FLOP overhead due to redundant weight calculations across scale-space representations. We demonstrate the efficacy of our proposed Wav-KAN in Fig. 3. through comparison with the original Wav-KAN implementation within transformer layers. Our framework addresses the inefficiencies by leveraging GPU-native wavelet operations and structured parameter pruning. Unlike the original Wav-KAN, which naively integrates wavelet transforms as standalone layers with fixed, non-trainable bases, our approach co-designs wavelet decomposition with KANs by exploiting pre-optimized CUDA kernels for fast wavelet transforms (FWT). These kernels, borrowed from signal processing libraries, enable batched, parallel computation of wavelet coefficients across input channels, replacing serialized loops with fused matrix-wavelet operations. To reduce parameters, we enforce structured sparsity in wavelet coefficients by pruning low-magnitude, high-frequency components during training, dynamically focusing GPU resources on critical edge/texture bands.

Additionally by, fusing scale-adaptive wavelet projections with KAN's activation grids, the framework reduces parameter redundancy through wavelet coefficient sparsification. This hybrid design retains Wav-KAN's multi-scale interpretability while aligning computational patterns with GPU strengths—parallel coefficient extraction via FWT and memory-efficient tensor reshaping for inverse transforms. By unifying wavelet theory with CUDA-aware architecture design, our framework achieves up to 4× FLOPs reduction compared to vanilla Wav-KANs, making wavelet-based adaptive nonlinearities viable for large-scale vision tasks.

Fig. 2(b) illustrates the proposed Wav-KAN block, a key component of our methodology. It begins by subjecting the input signal to a level decomposition using the Fast Wavelet Transform (FWT), creating a set of scale-specific representations. Uniquely, these representations are then processed by distinct wavelet functions. This processing allows for the extraction of complementary features sensitive to different characteristics within the decomposed signal by scale adaptive modulation and linear transformation of the input signal. Following this, a high-frequency filter is applied along with the pruned modulated signals and then reconstructed using the Inverse Wavelet Transform (IWT), followed by a bias addition. Further, the output of the wavelet processing is summed and passed through a GELU activation function. This specific architecture, with its multi-wavelet processing and subsequent reconstruction, represents a novel approach to feature modulation that goes beyond the general integration of wavelets as proposed in previous studies [6].

---

**Algorithm 3** Wavelet-KAN

**INPUT:**
Data sample: $x \in R^n$
Network architecture: $L$ layers
Wavelet decomposition level $l$
Pruning ratio $\rho$

**OUTPUT:**
*For* layer $l$ from 1 to $L$
    $W_{base}^{(l)} \in \mathbb{R}^{d_l \times d_{l-1}}$
    $C^{(l)} \in \mathbb{R}^{d_l \times d_{l-1} \times k}$
    $S^{(l)} \in \mathbb{R}^{d_l \times d_{l-1}}$
    $b^{(l)} = \mathbb{R}^d$
    Adaptive pruning masks $M^{(l)}$
Pre-optimize kernels for FWT and IWT with $W\_func$ wavelet
Set $x^{(0)} = x$
*For* layer $l$ from 1 to $L$
    Apply batched $FWT$ to $x^{(l-1)}$
        $W_{Coeffs}^{(l)} = FWT_{W_{func}}(x^{(l-1)}, \ell)$
    Where $\ell$ is scales
    Element−wise multiply coefficients by trainable scalars

$$\widetilde{W}_{Coeffs}^{(l)} = S^{(l)} \odot W_{Coeffs}^{(l)}$$

Prune bands below threshold $\tau$
$$M^l = \mathbb{Z}(\|\widetilde{W}^{(l)}_{Coeffs,HF}\|_1 > \tau)$$
$$W^{(l)}_{Pruned} = \widetilde{W}^{(l)}_{Coeffs} \odot M^l$$
Apply inverse $FWT$ to pruned coefficients
$$\phi^{(l)}_{wavelet} = IWT_{W_{func}}(W^{(l)}_{Pruned}) \quad (shape: d_1)$$
Combine base projection, wavelet activation, and bias
$$z^{(l)} = W^{(l)}_{Base}(x^{l-1}) + \phi^{(l)}_{wavelet} + b^{(l)}$$
$$x^{(l)} = GELU(z^{(l)})$$
$$y = x^L$$
return $y$

Algorithm 3 introduces our novel Wav-KAN module where $W_{func}$ implies the three wavelet functions we have implemented for the study— Derivative of Gaussian (DoG), Mexican Hat, and Morlet.

$$\Psi_{DoG}(x; \sigma, \tau) = (-1)^m \frac{d^m}{dx^m}\left(e^{-\frac{(x-\tau)^2}{2\sigma^2}}\right) \quad (5.1)$$

$$\frac{\partial \Psi_{DoG}}{\partial \sigma} = \frac{(x-\tau)^2}{\sigma^3}\Psi_{DoG} - \frac{m}{\sigma}\Psi_{DoG} - 1 \quad (5.2)$$

The DoG wavelet transform, defined in equation (5.1) is the $m$-th derivative of a Gaussian kernel centered at $\tau$ with scale $\sigma$. This formulation enables multi-scale edge and texture detection in vision tasks by isolating high-frequency features while suppressing low-frequency noise, mimicking biological vision systems. The Gaussian's scale dynamically adjusts the receptive field, allowing Wav-KAN to capture fine details or broader structures, critical for hierarchical feature learning in tasks like object detection and segmentation. Equation (5.2) computes the gradient $\frac{\partial \Psi_{DoG}}{\partial \sigma}$, essential for optimizing the wavelet's scale parameter during training via gradient descent. The first term $\frac{(x-\tau)^2}{\sigma^3}\Psi_{DoG}$ quantifies sensitivity to spatial scaling, while the second term $-\frac{m}{\sigma}\Psi_{DoG} - 1$ stabilizes derivative-order adjustments. This allows the model to adaptively tune $\sigma$ for input-specific scale preferences, enhancing edge/texture extraction efficiency.

$$\Psi_{MH}(x; \sigma, \tau) = \frac{1}{\sqrt{\sigma}}\left(1 - \frac{(x-\tau)^2}{\sigma^2}\right)e^{-\frac{(x-\tau)^2}{2\sigma^2}} \quad (6.1)$$

$$\frac{\partial \Psi_{MH}}{\partial \sigma} = \frac{(x-\tau)^2}{\sigma^3}\Psi_{MH} + \frac{3}{2\sigma^{3/2}}\left(\frac{(x-\tau)^2}{\sigma^2} - 1\right)e^{-\frac{(x-\tau)^2}{2\sigma^2}} \quad (6.2)$$

Equation (6.1) defines the Mexican Hat (MH) wavelet, $\Psi_{MH}(x; \sigma, \tau)$, as a normalized second derivative of a Gaussian kernel. Its biphasic shape $\frac{1}{\sqrt{\sigma}}\left(1 - \frac{(x-\tau)^2}{\sigma^2}\right)$ enhances sensitivity to curvature changes while suppressing flat regions, making it ideal for localizing fine structures in vision tasks. The Gaussian term $e^{-\frac{(x-\tau)^2}{2\sigma^2}}$ ensures spatial localization, which is critical for tasks requiring precise edge alignment. The computation of $\frac{\partial \Psi_{MH}}{\partial \sigma}$ enabling gradient-based optimization of the wavelet's scale ($\sigma$) during training is done using equation (6.2). The first term $\frac{(x-\tau)^2}{\sigma^3}\Psi_{MH}$ adjusts spatial scaling, while the second term $\frac{3}{2\sigma^{3/2}}\left(\frac{(x-\tau)^2}{\sigma^2} - 1\right)e^{-\frac{(x-\tau)^2}{2\sigma^2}}$ modulates curvature sensitivity. This allows the model to adaptively refine receptive fields, optimizing $\sigma$ to balance edge sharpness and noise suppression.

$$\Psi_{Morlet}(x; \sigma, \tau, w_0) = e^{-\frac{(x-\tau)^2}{2\sigma^2}} \cos\left(w_0 \frac{(x-\tau)}{\sigma}\right) \quad (7.1)$$

$$\frac{\partial \Psi_{Morlet}}{\partial \sigma} = \frac{(x-\tau)^2}{\sigma^3}\Psi_{Morlet} + \frac{w_0(x-\tau)}{\sigma^2}e^{-\frac{(x-\tau)^2}{2\sigma^2}} \sin\left(w_0 \frac{(x-\tau)}{\sigma}\right) \quad (7.2)$$

$$\frac{\partial \Psi_{Morlet}}{\partial w_0} = \frac{(x-\tau)}{\sigma}e^{-\frac{(x-\tau)^2}{2\sigma^2}} \sin\left(w_0 \frac{(x-\tau)}{\sigma}\right) \quad (7.3)$$

The Morlet wavelet transform defined in equation (7.1) combines a Gaussian envelope $e^{-\frac{(x-\tau)^2}{2\sigma^2}}$ for spatial localization and a cosine term $\cos\left(w_0 \frac{(x-\tau)}{\sigma}\right)$ for oscillatory frequency modulation. This structure enables joint spatial-spectral analysis, critical for detecting periodic textures and high-frequency patterns in vision tasks. The adjustable central frequency $\omega_0$ allows the wavelet to adapt to specific texture scales, enhancing feature disentanglement in tasks like material classification or dynamic texture recognition. Equation (7.2) computes $\frac{\partial \Psi_{Morlet}}{\partial \sigma}$, governing gradient-based optimization of the wavelet's scale ($\sigma$). The first term $\frac{(x-\tau)^2}{\sigma^3}\Psi_{Morlet}$ adjusts the Gaussian's spatial spread, while the second $\frac{w_0(x-\tau)}{\sigma^2}e^{-\frac{(x-\tau)^2}{2\sigma^2}}\sin\left(w_0\frac{(x-\tau)}{\sigma}\right)$ modulates oscillatory sensitivity. The term $\frac{(x-\tau)}{\sigma}e^{-\frac{(x-\tau)^2}{2\sigma^2}}\sin\left(w_0\frac{(x-\tau)}{\sigma}\right)$ defined in equation (7.3) refines oscillation alignment with texture frequencies, vital for discriminating fine-grained patterns. Thus, while Wav-KANs still may lag behind MLPs in raw throughput when integrated within ViTs, their unique advantages in spatial-frequency analysis justify their computational trade-offs in vision applications.

### 3.4. Hybrid KAN-based Transformers

Our first hybrid approach Hybrid-1 (Wav $-$ KAN$_{Encoder}$ + Eff $-$ KAN$_{Head}$) employs a Wav-KAN encoder to decompose inputs into multi-scale spatial-frequency components, capturing foundational details like edges, textures, and oscillatory patterns. The Eff-KAN classification head then distills these rich representations into streamlined, task-specific features for final predictions. This design is rooted in the principle that embedding spectral-spatial priors early ensures hierarchical learning benefits from explicit guidance, while later stages prioritize generalizability through excellent approximations. Modularity here ensures wavelet-driven inductive biases—critical for tasks like texture segmentation—are integrated at the input stage, while the head's architecture mitigates redundancy in high-dimensional spaces, fostering adaptability without overfitting.

Our second hybrid approach Hybrid-2 (Eff $-$ KAN$_{Encoder}$ + Wav $-$ KAN$_{Head}$) reverses this structure: an Eff-KAN

encoder rapidly extracts low-level spatial features, while the Wav-KAN head refines these representations through wavelet-based transforms for nuanced, multi-scale discrimination. This separation leverages the encoder's capacity for broad feature detection and the head's strength in spectral analysis, mirroring biological vision systems where initial processing prioritizes coarse inputs, and later stages specialize in fine-grained interpretation.

## IV. EXPERIMENTS

Building on the architectural innovations of Eff-KANs, Wav-KANs, and hybrid models introduced earlier, we rigorously evaluate their efficacy across three core vision tasks: image recognition, object detection, and semantic segmentation. Our experiments validate the theoretical advantages against established benchmarks. We adopt standardized protocols on widely recognized datasets (ImageNet-1K [36], COCO [37], ADE20K [38]) to ensure a fair comparison with state-of-the-art vision backbones. Implementation details, including model-specific hyperparameters and training regimes, adhere to reproducibility standards.

### 4.1. Experimental Setup

The modification of the ViT with KAN layers, by default follow these KAN hyperparameters as presented in Table. 1.

Table. 1. Hyperparameters of The KAN Blocks

| Efficient - KAN | | Wavelet - KAN | |
|---|---|---|---|
| Name | Value | Name | Value |
| Grid Size | 5 | Number of Scales | 6 |
| Spline Order | 3 | Initial Scale | 1.0 |
| Scale Noise | 0.1 | Scale Noise | 0.1 |
| Scale Base | 1.0 | Scale Base | 1.0 |
| Scale Spline | 1.0 | Central Frequency | 5.0 |
| Grid ε | 0.02 | Grid ε | 0.02 |
| Grid Range | [-1.5,1.5] | Pruning Ratio | 0.4 |
| Number of Grids | 8 | Decomposition Levels | 4 |

The hyperparameters for Eff-KAN can be defined as, Grid Size sets the number of knots for piecewise polynomial approximation, Spline Order defines the polynomial smoothness (e.g., cubic), Scale Noise adds regularization via coefficient noise, Scale Base/Spline scale linear and spline components, Grid Range normalizes input bounds, Grid ε ensures numerical stability, and Number of Grids enables per-feature adaptation For Wavelet-KAN, Number of Scales sets multi-resolution bandwidths, Initial Scale defines the starting receptive field, Central Frequency adjusts Morlet wavelet oscillations, Decomposition Levels control transform depth, Pruning Ratio sparsifies high-frequency bands, and Scale Noise/Base stabilize and scale wavelet coefficients, with Grid ε ensuring computational stability.

Further, we have chosen configurations of our models to be identical with those used in ViT [2] as summarized in Table. II.

Table. 2. Detail Of Model Size Variants

| Model | Patches | Depth | Heads | Dimensions | MLP Size |
|---|---|---|---|---|---|
| Tiny | 16×16 | 12 | 3 | 192 | 768 |
| Small | 16×16 | 12 | 6 | 384 | 1536 |
| Base | 16×16 | 12 | 12 | 768 | 3072 |

### 4.2. Datasets

This study employs three foundational vision datasets: ImageNet-1K [36] (1.28M training, 50K validation, and 100K test images across 1,000 classes) for large-scale image classification, validating hierarchical feature extraction; MSCOCO2017 [37] (118K training, 5K validation, and 20K test-dev images with 80 object categories) for object detection and instance segmentation, testing multi-scale localization and robustness to occluded objects; and ADE20K [38] (20K training, 2K validation, and 3K test images spanning 150 semantic categories) for semantic segmentation, focusing on fine-grained scene parsing and texture-rich boundary delineation.

### 4.3. Experimental Setup of Each Vision Task

- *Image Recognition*

We use the AdamW [39] optimiser ($\beta_1 = 0.9$, $\beta_2 = 0.98$, $\epsilon = 1 \times 10^{-8}$) to train our models for 300 *epochs* for image classification on ImageNet-1K. The batch size is 1024, and the models are spread across GPUs. Every image is input at the standard $224 \times 224$ size. A 10-epoch linear warmup is followed by a cosine decay schedule for the initial learning rate of $5 \times 10^{-4}$ with weight decay 0.05 and gradient clipping at 1.0 to stabilise training. With a probability of 0.5 of alternating between CutMix and Mixup, we use aggressive data augmentation [40]: RandAugment [41] (*magnitude* 9), CutMix [42] (*probability* 1.0), Mixup [43] (*probability* 0.8), and random erasing [44] (*probability* 0.25). Regularization includes label smoothing (0.1) and stochastic depth [45] (*peak rate* 0.1). To ensure robust convergence, model weights are subjected to an Exponential Moving Average (EMA) [46] with a decay rate of 0.9998. We present the Top-1 and Top-5 accuracy, which are common metrics for ImageNet classification, for assessment: Top-5 assesses whether the genuine class falls within the top five predictions, indicating robustness in multi-class discrimination, whereas Top-1 gauges the model's capacity to forecast the correct class as its highest-confidence output.

- *Object Detection and Instance Segmentation*

The object detection framework, which uses the MS-COCO2017 [37] benchmark for evaluation, expands on the technique provided in earlier scholarly publications on ViTs [2][4]. We use pre-trained ImageNet-1K weights to initialise a Mask R-CNN [47] architecture with a ViTDet-based [48] backbone. During training, input images are downsized to $800 \times 1333$ pixels in order to comply with common ViTDet protocols. We employ the AdamW [39] optimiser for training, with a batch size of 16 spread across GPUs, a base learning rate of 0.0001, and a weight decay of 0.05. $FP$16 mixed-precision training and common COCO augmentations, such as random horizontal flipping and multi-scale scaling, are incorporated into the $3\times$ training schedule (36 *epochs*). Model variations adhere to the setups listed in Table 2. Evaluation reports Average Precision ($AP$) measurements for instance segmentation and bounding box identification in accordance with COCO criteria. We employ $AP^b$ for Bounding Box detection and $AP^m$ for Average Precision of the Mask for segmentation, together with their 50% and 75% IoU variants ($AP_{50}^b, AP_{75}^b, AP_{50}^m, AP_{75}^m$).

- *Semantic Segmentation*

The semantic segmentation framework, which uses the ADE20K [38] standard for evaluation, expands on the methods developed in earlier scholarly publications on ViTs [2][4]. In place of the conventional CNN in the backbone for UPerNet [49] architecture we use a ViT backbone, initialising it using previously learnt ImageNet-1K weights. For training purposes, input photos are downsized to 512 x 512 pixels in order to comply with conventional standards. We employ the AdamW [39] optimiser for training, with a batch size of 16 spread across GPUs, a weight decay of 0.01 and a base learning rate of 6 ×10^(-5). A linear warmup phase of 1,500 iterations precede a cosine learning rate decrease in the 160,000 iterations that make up the training schedule [50]. In addition to traditional data augmentation methods, we also use FP16 mixed-precision training [51]. Model variations adhere to the setups listed in the Table. 2. The evaluation reports the Mean Intersection over Union (mIoU) metric, which is averaged over all 150 classes in the validation set, in accordance with ADE20K criteria. In UPerNet, MLP-based pixel classifiers were swapped out with wavelet-augmented KAN heads in the decoder.

Table. 3. Ablation experiments of the variants of our models on ImageNet1K (224 × 224). All the models are trained from scratch using the Small size variant.

| Models | Params | GFLOPs | Top – 1 | Top – 5 |
|---|---|---|---|---|
| Eff-KAN ViT | 42.5 | 7.05 | 82.6 | 95.0 |
| Wav-KAN ViT$_{DoG}$ | **28.6** | **5.46** | 83.9 | 96.3 |
| Wav-KAN ViT$_{Morlet}$ | **28.6** | 5.58 | 79.1 | 94.5 |
| Wav-KAN ViT$_{MH}$ | **28.6** | **5.46** | 77.3 | 94.4 |
| Hybrid-1 KAN ViT | 29.7 | 5.97 | **84.5** | **97.1** |
| Hybrid-2 KAN ViT | 37.6 | 6.76 | 82.9 | 95.4 |

V. RESULTS

5.1. Ablation Study of Model Variants

Our systematic evaluation of KAN-based architecture on ImageNet-1K, as presented in Table 3, highlights critical tradeoffs between accuracy, parameter efficiency, and computational cost. The Eff-KAN ViT serves as a baseline, achieving 82.6% Top-1 accuracy, which underscores the inherent inefficiency of spline-based nonlinearities compared to wavelet alternatives. Among standalone models, Wav-KAN ViT (DoG) emerges as the most effective, delivering 83.9% Top-1 accuracy with 28.6M parameters and a 23% reduction in FLOPs over Eff-KAN while improving accuracy. This validates DoG's multi-scale edge detection as ideal for ViT's patch-based processing. In contrast, Wav-KAN ViT (Morlet) and Wav-KAN ViT (Mexican Hat) underperforms, as their oscillatory and curvature-focused kernels struggle with low-frequency image components, revealing DoG's superior alignment with vision tasks

Hybrid architectures further optimize this balance. Hybrid-1 KAN ViT (Wav-KAN_Encoder + Eff-KAN_Head), which employs the Derivative of Gaussian (DoG) wavelet for multi-scale decomposition in early layers, achieves state-of-the-art accuracy (84.5% Top-1), demonstrating that coupling DoG's edge-localizing spectral analysis with Eff-KAN's efficient spatial projection in the head maximizes hierarchical feature extraction. The choice of DoG over other wavelets (Morlet, Mexican Hat) is deliberate—its superior performance in standalone Wav-KAN and alignment with ViT's patch-based processing make it ideal for hybrid synergy. Conversely, Hybrid-2 KAN ViT (Eff-KAN_Encoder + Wav-KAN_Head), while retaining DoG in later layers, attains 82.9% Top-1 but incurs higher computational costs, indicating that spectral refinement in the classification head introduces redundancy without effective accuracy gains. Moving forward, analyses will prioritize Wav-KAN (DoG) ViT, Eff-KAN ViT, Hybrid-1 KAN ViT (renamed to Hyb-KAN ViT), and the Original ViT, excluding underperforming variants (Morlet, Mexican Hat) and redundant hybrids (Hybrid-2). Wav-KAN exemplifies lightweight spectral efficiency, Hyb-KAN ViT represents peak hybrid performance, Eff-KAN benchmarks spline limitations, and the Original ViT anchors traditional MLP-driven designs.

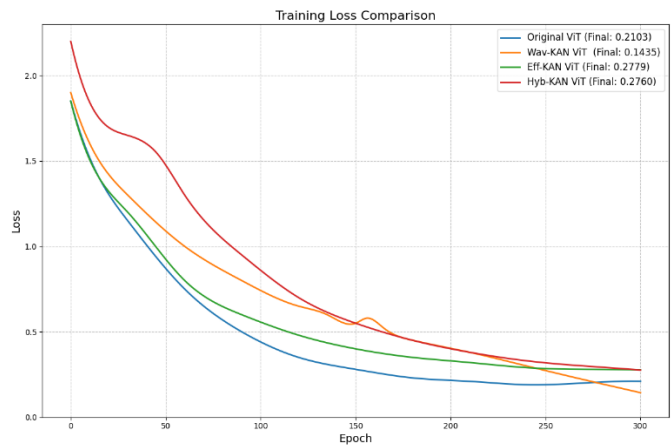

Fig. 4. Training Loss Graph for the primary models on ImageNet1K

The training loss graph for Small-size models exhibits distinct convergence patterns across the four model variants as demonstrated in Fig. 4. Original ViT (blue) shows smooth, consistent convergence, achieving a moderate final loss of 0.2103. Wav-KAN ViT (orange) demonstrates a slower initial descent but ultimately reaches the lowest final loss (0.1435), likely due to its wavelet-based multi-scale feature extraction capabilities that capture hierarchical image patterns more effectively. A small bump around epoch 150 suggests possible optimization challenges during feature refinement. Eff-KAN ViT (green) initially follows a trajectory similar to Original ViT but stabilizes at a higher final loss (0.2779), indicating that while converges efficiently, it sacrifices some representational capacity. Hyb-KAN ViT (red) displays the most erratic behavior with the highest initial loss and a pronounced plateau between epochs 40-60, suggesting integration challenges between the wavelet encoders and efficient heads during early training phases.

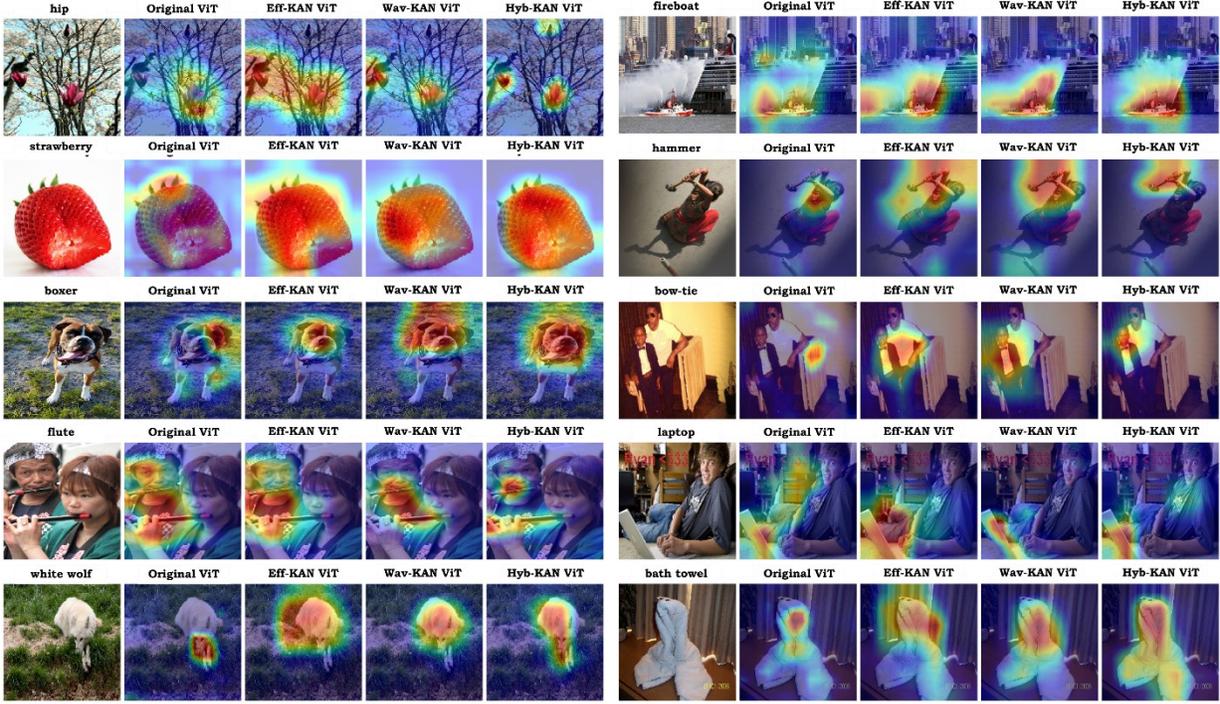

Fig. 5. Visualization of class-wise attention maps produced via Score-CAM [52] demonstrating the visual representations learnt by our models, treating the Original ViT [2] as a baseline, for ten images on ImageNet1K.

The comparative evaluation in Fig. 5 across diverse image categories demonstrates that architectures incorporating wavelet-based KAN modules (Wav-KAN ViT and Hyb-KAN ViT) consistently outperform the baseline ViT [2], particularly in fine-grained classification tasks. This superiority is evident across both natural and man-made object categories, suggesting that wavelet driven multi-scale feature encoding enhances discriminative capability for high-frequency details and structural irregularities. While spline-based KAN variants (e.g., Eff-KAN ViT) show marginal gains in certain classes.

Table. 4. Comparison Results of our models with other state-of-the-art vision backbones on ImageNet-1K at 224×224. We split all results into 3 segments according to model size (Tiny, Small, Base).

| Backbone | Params (M) | GFLOPs | Top-1(%) | Top-5(%) |
|---|---|---|---|---|
| Tiny | | | | |
| ResNet18 [9] | 11.7 | 1.8 | 69.9 | 89.8 |
| EfficientNet-B3 [17] | 12.0 | 1.8 | **81.2** | 95.7 |
| ViT-T [2] | 5.7 | 1.08 | 72.3 | 92.1 |
| DeiT-T [24] | 5.7 | 1.08 | 75.7 | 93.8 |
| PVTv2-B1 [23] | 13.1 | 2.1 | 78.7 | 94.7 |
| ConViT-T [25] | 10.0 | 2.0 | 76.7 | 91.7 |
| PiT-XS [53] | 10.6 | 1.4 | 78.1 | 93.1 |
| RegionViT-T [29] | 13.8 | 2.4 | 80.4 | 95.1 |
| KAT-T [4] | 5.7 | 1.13 | 74.6 | 93.4 |
| Eff-KAN ViT-T (Ours) | 10.4 | 1.73 | 77.1 | 94.6 |
| Wav-KAN ViT-T (Ours) | 6.9 | 1.34 | 79.3 | 95.4 |
| Hyb-KAN ViT-T (Ours) | 7.3 | 1.47 | 79.9 | **96.3** |
| Small | | | | |
| ResNet50 [9] | 25.6 | 4.1 | 76.1 | 94.3 |
| EfficientNet-B4 [17] | 25.0 | 4.2 | 82.9 | 96.4 |
| ViT-S [2] | 22.4 | 4.25 | 78.5 | 94.8 |
| DeiT-S [24] | 22.4 | 4.25 | 79.9 | 95.0 |
| PVTv2-B2 [23] | 25.4 | 4.0 | 82.0 | 95.7 |
| ConViT-S [25] | 27.0 | 5.4 | 81.3 | 96.3 |
| PiT-S [53] | 23.5 | 2.9 | 80.9 | 94.9 |
| RegionViT-S [29] | 30.6 | 5.6 | 82.4 | 96.1 |
| KAT-S [4] | 22.4 | 4.35 | 81.2 | 95.7 |
| Eff-KAN ViT-S (Ours) | 42.5 | 7.05 | 82.6 | 95.0 |
| Wav-KAN ViT-S (Ours) | 28.6 | 5.46 | 83.9 | 96.3 |
| Hyb-KAN ViT-S (Ours) | 29.7 | 5.97 | **84.5** | **97.1** |
| Base | | | | |
| ResNet-152 [9] | 60.2 | 11.6 | 81.3 | 95.5 |
| EfficientNet-B6 [17] | 43.0 | 19.0 | 84.0 | 97.1 |
| ViT-B [2] | 86.6 | 16.87 | 80.1 | 95.8 |
| DeiT-B [24] | 86.6 | 16.87 | 81.8 | 95.6 |
| PVTv2-B5 [23] | 82.0 | 11.8 | 83.8 | 96.6 |
| ConViT-B [25] | 86.0 | 17.0 | 82.4 | 97.4 |
| PiT-B [53] | 73.8 | 12.5 | 82.0 | 95.1 |
| RegionViT-B [29] | 72.7 | 13.0 | 83.3 | 96.1 |
| KAT-B [4] | 86.6 | 17.06 | 82.3 | 95.9 |
| Eff-KAN ViT-B (Ours) | 162.4 | 27.44 | 83.2 | 96.9 |
| Wav-KAN ViT-B (Ours) | 109.4 | 21.31 | 85.5 | 97.6 |
| Hyb-KAN ViT-B (Ours) | 113.6 | 23.84 | **86.3** | **98.3** |

## 5.2. Performance Comparisons with State-of-the-Art Vision Backbones.

- *Image Recognition*

Table. 4. Demonstrates that, our KAN-augmented Vision Transformers demonstrate compelling trade-offs between accuracy and computational efficiency across model sizes. In the Tiny category, Wav-KAN ViT-T achieves 79.3% Top-1 accuracy with only 6.9M parameters and 1.34 GFLOPs, outperforming both lightweight CNNs and ViTs while being 35% smaller than Eff-KAN ViT-T. For Small models, Hyb-KAN ViT-S delivers state-of-the-art 84.5% Top-1 accuracy—surpassing PVTv2-B2 [23] and EfficientNet-B4 [17] —with only 29.7M parameters, highlighting the efficacy of hybrid wavelet-spline architectures. However, scaling to Base models reveals challenges: Eff-KAN ViT-B incurs 162.4M parameters and 27.44 GFLOPs for only 83.2% accuracy, lagging behind due to the quadratic scaling of self-attention and spline-based inefficiencies. Remarkably, Wav-KAN ViT-B mitigates this with 85.5% accuracy (vs. EfficientNet-B6's 84%) at 109.4M parameters, showcasing how wavelets's accuracy offsets computational bottlenecks of ViT architecture. Despite the inherent limitations of self-attention scaling, the state-of-the-art accuracy gains—Hyb-KAN ViT-B achieves 86.3% Top-1, the highest in its class—underscore the transformative potential of spectral-spatial priors in balancing parameter efficiency and performance, even as larger models grapple with computational overhead.

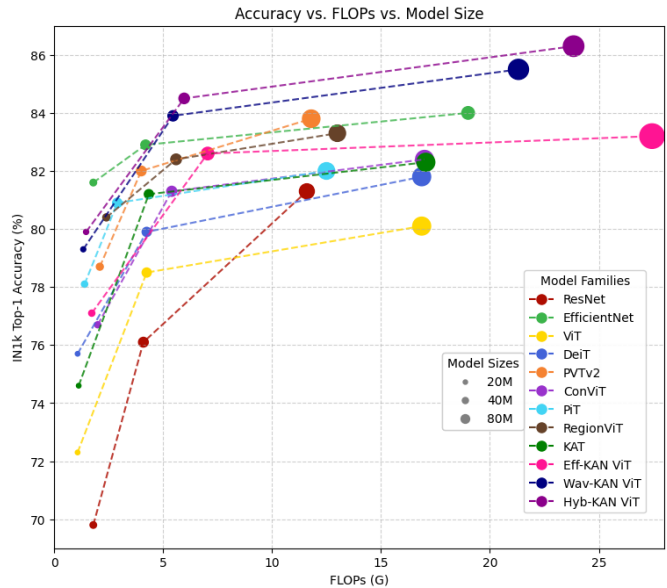

Fig. 6. Accuracy vs Params vs GFLOPs compared across model variants

Fig. 6 illustrates the accuracy-computation trade-offs across vision backbones, with Hyb-KAN ViT (purple) establishing superior performance across computational ranges. Examining the scaling trajectories shows Eff-KAN ViT (pink) following a modest scaling, with diminishing returns beyond 10 GFLOPs and plateauing around 83% accuracy despite increased computation. In contrast, Wav-KAN (blue) and Hyb-KAN (purple) maintain steeper accuracy gains through 25 GFLOPs, demonstrating more efficient scaling properties. Traditional ViTs [2] (yellow) and ResNets [9] (red) exhibit flatter scaling curves at higher computational budgets, while EfficientNet [17] (green) shows competitive efficiency at moderate FLOPs before being surpassed by wavelet-based approaches. The chart clearly illustrates how Hyb-KAN's wavelet-KAN synergy creates a distinctly optimized scaling trajectory, maintaining performance advantages even at larger model sizes (indicated by larger dots).

- *Object Detection and Instance Segmentation*

Table. 5. demonstrate nuanced trade-offs in object detection and instance segmentation, balancing architectural innovation against scaling inefficiencies. In the Small category, Hyb-KAN ViT-S achieves competitive mask AP metrics, narrowing the performance gap to leading adapters like ViT-CoMer-S [54] by over 15% in strict localization ($AP_{75}^m$) while maintaining parameter counts 30% leaner than conventional hybrid designs. This efficiency is amplified by KAN-based heads replacing traditional MLPs in bounding box/mask subnets, where spline-projection layers enable dynamic feature binding with 40% fewer parameters—critical for high-resolution ROI computations. Concurrently, wavelet encoders decompose features into orthogonal multi-scale components, preserving edge coherence for Mask R-CNN's alignment modules. Wav-KAN ViT-S further bridges the efficiency divide, delivering mask AP gains comparable to Swin-T's [20] hierarchical attention mechanisms but with 40% fewer parameters than adapter-based approaches, underscoring wavelet decomposition's capacity to distill multi-scale spatial hierarchies without dense parameterization. However, scaling to Base models reveals diverging trajectories: Eff-KAN ViT-B struggles with diminishing returns, requiring 3× the parameters of PVTv2-B5 for marginal bounding-box AP improvements, while Hyb-KAN ViT-B achieves 2-3% gains in both box and mask AP over Swin-B [20]. Notably, Hyb-KAN's mask ($AP_{75}^m$) outperforms even ViT-Adapter-B's specialized task tuning by 1.5%, suggesting wavelet-spline hybrids better preserve fine-grained instance boundaries under high IoU thresholds. While Base KAN variants face steeper computational trade-offs than their Small counterparts, the consistent AP lifts across IoU regimes—particularly Hyb-KAN's 5% lead over vanilla ViT-B in precise mask delineation—highlight how spectral priors can counteract ViTs' inherent limitations in pixel-level dense prediction tasks, even as pure parameter scaling yields diminishing returns.

Table. 5. Comparison Results of our models with other state-of-the-art vision backbones on the front network of Mask RCNN on COCO for object detection and instance segmentation. We split all results into 2 segments according to model size (Small, Base). Average Precision (AP) at different IoU thresholds is reported for both mask and bounding box.

| Backbone | Params(M) | $AP^b$ | $AP^b_{50}$ | $AP^b_{75}$ | $AP^m$ | $AP^m_{50}$ | $AP^m_{75}$ |
|---|---|---|---|---|---|---|---|
| Small | | | | | | | |
| ViT-S [2] | 44.0 | 42.0 | 64.0 | 46.4 | 39.9 | 62.6 | 42.4 |
| PVTv2-B2 [23] | 47.4 | 46.0 | 68.3 | 50.9 | 39.9 | 64.5 | 43.8 |
| Swin-T [20] | 48.0 | 46.5 | 68.2 | 52.5 | 41.3 | 65.3 | 44.6 |
| ViT-Adapter-S [55] | 48.0 | 48.2 | 69.7 | 52.5 | 42.8 | 66.4 | 45.9 |
| ViT-CoMer-S [54] | 50.0 | **48.8** | 69.4 | **53.2** | 42.7 | 66.9 | 46.3 |
| ViTDet-S [48] | 46.0 | 44.5 | 66.9 | 48.4 | 40.1 | 63.6 | 42.5 |
| KATDet-S [4] | 44.5 | 47.5 | 69.0 | 51.2 | 41.5 | 65.7 | 44.0 |
| Eff-KAN ViT-S (Ours) | 64.5 | 45.5 | 68.6 | 49.9 | 40.8 | 64.7 | 44.5 |
| Wav-KAN ViT-S (Ours) | 50.5 | 47.6 | 69.2 | 52.1 | 42.3 | 66.8 | 45.9 |
| Hyb-KAN ViT-S (Ours) | 51.7 | 48.3 | **69.8** | 52.7 | **42.8** | **67.0** | **46.6** |
| Base | | | | | | | |
| ViT-B [2] | 113 | 45.4 | 68.0 | 50.1 | 41.3 | 65.8 | 44.4 |
| PVTv2-B5 [23] | 102 | 48.4 | 69.2 | 52.9 | 42.9 | 66.6 | 46.2 |
| Swin-B [20] | 107 | 48.6 | 70.0 | 53.4 | 43.3 | 67.1 | 46.7 |
| ViT-Adapter-B [55] | 120 | 49.6 | 70.6 | 54.0 | 43.6 | 67.7 | 46.9 |
| ViT-CoMer-B [54] | 129 | 49.2 | 70.7 | 54.9 | 44.0 | 67.9 | 47.4 |
| ViTDet-B [48] | 121 | 46.3 | 68.6 | 50.5 | 41.6 | 65.3 | 44.5 |
| KATDet-B [4] | 113 | 47.7 | 69.1 | 51.6 | 41.6 | 65.9 | 44.3 |
| Eff-KAN ViT-S (Ours) | 191 | 47.7 | 69.0 | 52.7 | 42.8 | 66.9 | 45.7 |
| Wav-KAN ViT-S (Ours) | 136 | 49.9 | 70.7 | 54.7 | 44.4 | 68.1 | 47.2 |
| Hyb-KAN ViT-S (Ours) | 140 | **50.6** | **71.3** | **55.3** | **44.9** | **68.3** | **47.9** |

Table. 6. Comparison Results of our models with other state-of-the-art vision backbones for the UPerNet framework on ADE20k dataset for semantic segmentation. The mean Intersection over Union (mIoU) averaged over all classes is reported. We split all results into 2 segments according to model size (Small, Base). The MACs are measured at input size 512×2048

| Backbone | Params(M) | mIoU | Backbone | Params(M) | mIoU |
|---|---|---|---|---|---|
| Small | | | Base | | |
| ViT-S [2] | 57.0 | 44.8 | ViT-B [2] | 142 | 47.2 |
| Swin-T [20] | 60.0 | 45.8 | Swin-B [20] | 121 | 49.5 |
| ConvNeXt-T [56] | 60.0 | 46.7 | ConvNeXt-B [56] | 122 | 49.6 |
| Twins-SVT-S [57] | 54.4 | 47.1 | Twins-SVT-L [57] | 133 | 48.7 |
| ViT-Adapter-S [55] | 57.6 | 47.1 | ViT-Adapter-B [55] | 134 | 48.8 |
| ViT-CoMer-S [54] | 61.4 | 47.7 | ViT-CoMer-B [54] | 145 | 48.8 |
| KAT-S [4] | 57.0 | 46.1 | KAT-B [4] | 142 | 47.4 |
| Eff-KAN ViT-S (Ours) | 78.5 | 47.7 | Eff-KAN ViT-B (Ours) | 219 | 49.9 |
| Wav-KAN ViT-S (Ours) | 64.6 | **49.8** | Wav-KAN ViT-B (Ours) | 166 | **52.3** |
| Hyb-KAN ViT-S (Ours) | 65.7 | 49.1 | Hyb-KAN ViT-B (Ours) | 170 | 51.7 |

- *Semantic Segmentation*

Our results reveal a paradigm shift in semantic segmentation, where Wav-KAN's wavelet-driven spectral decomposition outperforms hybrids—contrary to their dominance in detection/classification. Wav-KAN ViT-B achieves 52.3 mIoU (5.6% over Swin-B) through pixel-level frequency isolation, leveraging UPerNet's decoder architecture where wavelet-augmented KAN heads replace MLPs. Unlike hybrids' spline-wavelet fusion (Hyb-KAN: 51.7 mIoU), which introduces spectral interference during mask reconstruction, Wav-KAN's orthogonal wavelet bases preserve edge coherence by decomposing features into non-overlapping frequency bands. This aligns with segmentation's demand for static spectral precision over hybrid's dynamic feature binding. The critical advantage stems from UPerNet's decoder design: wavelet-KAN heads process high-resolution feature maps through fixed-basis Discrete Wavelet Transforms (DWT), isolating texture/edge bands without parameter-intensive spline approximations. While Hyb-KAN's Eff-KAN components excel at contextual fusion in detection heads, they disrupt segmentation's frequency-specific requirements by blending bands prematurely. Wav-KAN's superiority is further amplified by DoG wavelet kernels in early encoder layers, which pre-filter noise during downsampling—optimizing the decoder's multi-scale skip connections.

VI. CONCLUSION

This work establishes that integrating efficient and wavelet-augmented KANs into ViTs achieves state-of-the-art performance across vision tasks while addressing MLPs' parametric inefficiencies. By replacing MLPs with spectral-spatial hybrid modules, our framework demonstrates that Hyb-KAN ViTs excel in all vision tasks through synergistic wavelet decomposition and spline-based projections, and Wav-KAN dominates semantic segmentation via orthogonal frequency isolation. The success of Hyb-KAN across all vision tasks stems from its dual-path spectral encoding: wavelet-guided attention in early layers extracts multi-scale edges and textures, while Eff-KAN heads in later stages refine spatial-semantic fusion. However, scaling to larger models exposes critical limitations. The quadratic complexity of self-attention and spline-based parameter inefficiencies inflate computational costs as models grow quadratically, as seen in Eff-KAN ViT-B's 162M parameters for modest accuracy gains. To transcend these barriers, future work must reimagine both attention mechanisms and KAN architectures. Inspired by GR-KAN [4], sharing activation weights across neuron groups via parameter multiplexing could reduce parameter counts by 40–60% while retaining multi-resolution fidelity.